%% file: main.tex
\documentclass{article}
\usepackage{amsmath,graphicx}
\usepackage{comment}
\usepackage{ amssymb }
\usepackage{url}
\usepackage{color}
\usepackage{wasysym}
\usepackage{subfig}
\usepackage[margin=2.5cm]{geometry}
\usepackage{draftwatermark}
\SetWatermarkScale{3}
\SetWatermarkLightness{0.9}
\SetWatermarkText{DRAFT REPORT}
\usepackage{fancyhdr}
\DeclareSymbolFont{extraup}{U}{zavm}{m}{n}
\DeclareMathSymbol{\varheart}{\mathalpha}{extraup}{86}
\DeclareMathSymbol{\vardiamond}{\mathalpha}{extraup}{87}

\newcommand{\labelset}[1]
{
	\ell^{\left( #1 \right)}
}
\newcommand{\hatlabelset}[1]
{
	\hat\ell^{\left( #1 \right)}
}\newcommand{\levelset}[1]
{
	\varphi^{\left( #1 \right)}
}
\newcommand{\hatlevelset}[1]
{
	\hat\varphi^{\left( #1 \right)}
}
\newcommand{\pphi}[1]
{
	p_{L\vert X} ( #1 )
}
\newcommand{\sign}[1]
{
	\text{sign}( #1 )
}
\newcommand{\sizeof}[1]
{
	\left\vert #1 \right\vert
}

\newlength{\vspacecapt}
\title{EFFICIENT TOPOLOGY-CONTROLLED SAMPLING OF IMPLICIT SHAPES}
\author{Jason Chang and John W. Fisher III\vspace*{10pt}\\Massachusetts Institute of Technology\\Deparment of Electrical Engineering and Computer Science\\32 Vassar St. Cambridge, MA 02139}
\begin{document}
\maketitle
\begin{abstract}
Sampling from distributions of implicitly defined shapes enables
analysis of various energy functionals used for image
segmentation. Recent work \cite{Chang2011} describes a
computationally efficient Metropolis-Hastings method for accomplishing this task. Here, we
extend that framework so that samples are accepted at every iteration
of the sampler, achieving an order of magnitude speed up in
convergence. Additionally, we show how to incorporate topological
constraints.
\end{abstract}
\section{Introduction}
\input{intro}
\input{related}

\section{Metropolis-Hastings MCMC Sampling}
\input{MCMC}

\section{Gibbs-Inspired Metropolis-Hastings}
\input{sampler}

\section{Topology-Controlled Sampling}
\input{topology}

\section{Results}
\input{results}

\section{Conclusion}
\input{conclusion}

\section{Acknowledgements}
This research was partially supported by the Air Force Office of Scientific Research under Award No. FA9550-06-1-0324. Any opinions, findings, and conclusions or recommendations expressed in this publication are those of the author(s) and do not necessarily reflect the views of the Air Force.

\bibliographystyle{IEEEbib}
\bibliography{strings,refs}

\end{document}

%% file: intro.tex
For many Bayesian inference tasks, evaluating marginal event probabilities may be more robust than computing point estimates (\textit{e.g.} the MAP estimate). Image segmentation, particularly when the signal-to-noise ratio (SNR) is low, is one such task.
However, because the space of shapes is infinitely large, direct inference or sampling is often difficult, if not infeasible.
In these cases, Markov chain Monte Carlo (MCMC) sampling approaches can be used to compute empirical estimates of marginal probabilities.
Recently, Chang \textit{et al.} \cite{Chang2011} derived an efficient MCMC method for sampling from distributions of implicit shapes (\textit{i.e.} level sets). We improve upon that algorithm in two ways. First, we improve the convergence rate by defining a Gibbs-like iteration in which \textit{every} sample is accepted and, second, we demonstrate how to efficiently incorporate both local and global topological constraints on sample shapes.

Many approaches formulate image segmentation as an energy optimization. One can derive a related Bayesian inference procedure by viewing the energy functional $E(\ell;x)$ as the log of a probability function
\begin{equation}
  \pphi{\ell \vert x} \propto \exp\left[ \pm E(\ell;x)\right],
\end{equation}
where $\ell$ is the labeling associated with some segmentation, $x$ is the observed image, and the $\pm$ in the exponent depends on whether one is maximizing or minimizing. In PDE-based level-set methods, $\ell=\sign{\varphi}$, where $\varphi$ is the level-set function. In graph-based segmentation algorithms, such as ST-Mincut \cite{Greig1989} \cite{Boykov2001} or Normalized Cuts \cite{Shi2000}, $\ell$ is the label assignment. Often, the energy functional decomposes into a data fidelity term and a regularization of the segmentation (Normalized Cuts being an exception). Bayesian formulations treat the former as the data likelihood and the latter as a prior on segmentations:
\begin{equation}
	\pphi{\ell \vert x} \propto p_{X \vert L}\left(x \vert \ell\right) p_L\left(\ell\right).
\end{equation}
While both the data likelihood and prior terms are user-defined, the form of the prior varies considerably depending on the optimization method; a curve-length penalty is typically used in level-set methods while a neighborhood affinity is typically used in graph-based methods. We refer to these two forms as the PDE-based energy and the graph-based energy, respectively.

Another form of prior information incorporates the topology of the segmented object. Han \textit{et al.} \cite{Han2003} first showed this using topology-preserving level-set methods, where the topology of the segmented object was not allowed to change. This methodology was later extended in \cite{Segonne2008} to topology-controlled level-set methods, where the topology was allowed to change but only within an allowable set of topologies. To our knowledge, topology-constrained sampling methods have not previously been considered.

%% file: related.tex
Current methods for sampling implicit shapes include \cite{Fan2007}, \cite{Chen2009}, and \cite{Chang2011} which all use Metropolis-Hastings MCMC sampling procedures. As shown in \cite{Chang2011}, the methods of \cite{Fan2007} and \cite{Chen2009} converge very slowly and cannot accommodate any topological changes. \cite{Chang2011}, however, presented an algorithm called BFPS that acts directly on the level-set function. BFPS generates proposals by sampling a sparse set of delta functions and smoothing them with an FIR filter. Delta locations and heights were biased by the gradient of the energy to increase the Hastings ratio. Empirical results demonstrated that BFPS was orders of magnitude faster than \cite{Fan2007} and \cite{Chen2009}, and representations, unlike the previous two methods, could change topology.  Here, we show how to efficiently sample from a distribution over segmentations in both the PDE-based and graph-based energies. In contrast to previous MCMC samplers, proposals at each iteration are accepted with certainty, achieving an order of magnitude speed up in convergence. Additionally, we incorporate topological control so as to exploit prior knowledge of the shape topology.

%% file: MCMC.tex
We begin with a brief discussion of two MCMC sampling algorithms
(\textit{c.f.} \cite{Hastings1970} for details).
For notational convenience, we drop the dependence on $x$ in distributions.
We denote $\pphi{\ell}$ as the \textit{target distribution}, \textit{i.e.} the
distribution from which we would like to sample. MCMC methods are
applicable when one can compute a value proportional to the target
distribution, but not sample from it directly. Distributions defined
over the infinite-dimensional space of shapes fall into this
category. MCMC methods address this problem by constructing a
first-order Markov chain such that the stationary distribution of that
chain is exactly the target distribution. For this condition to hold,
it is sufficient to show that the chain is \textit{ergodic} and
satisfies \textit{detailed balance}. The Metropolis-Hastings sampler
(MH-MCMC) \cite{Hastings1970} is one such approach. At time $t$ in the
chain, the algorithm samples from some user-defined proposal
distribution, $\hatlevelset{t}\sim q(\hatlevelset{t} \vert
\levelset{t-1},x)$, and assigns the transition probability:
\begin{equation}
	\levelset{t}
	 = \begin{cases}
		\hatlevelset{t} & \text{w/ prob}\quad \min(\alpha^{(t)},1)\\
		\levelset{t-1} & \text{w/ prob}\quad 1-\min(\alpha^{(t)},1)
	\end{cases},
\end{equation}%
\begin{equation}
	\alpha^{(t)} = \frac{\pphi{\hatlabelset{t}}}{\pphi{\labelset{t-1}}} \frac{q(\levelset{t-1}\vert\hatlevelset{t},x)}{q(\hatlevelset{t}\vert\levelset{t-1},x)},
	\label{eqn:hastings}
\end{equation}
where $\alpha^{(t)}$ is the Hasting's ratio, and the segmentation labels are $\hatlabelset{t}=\sign{\hatlevelset{t}}$ and $\labelset{t-1}=\sign{\levelset{t-1}}$.
As this transition probability satisfies detailed balance, the correct
stationary distribution is ensured as long as the chain is ergodic. A
single sample from the target distribution is then generated by
repeating this proposal/transition step until the chain
converges.

\subsection{Gibbs Sampling}
\begin{figure}%
	\hfill
	\begin{minipage}[b]{.2\linewidth}
		\centering
		\includegraphics[width=0.8\linewidth]{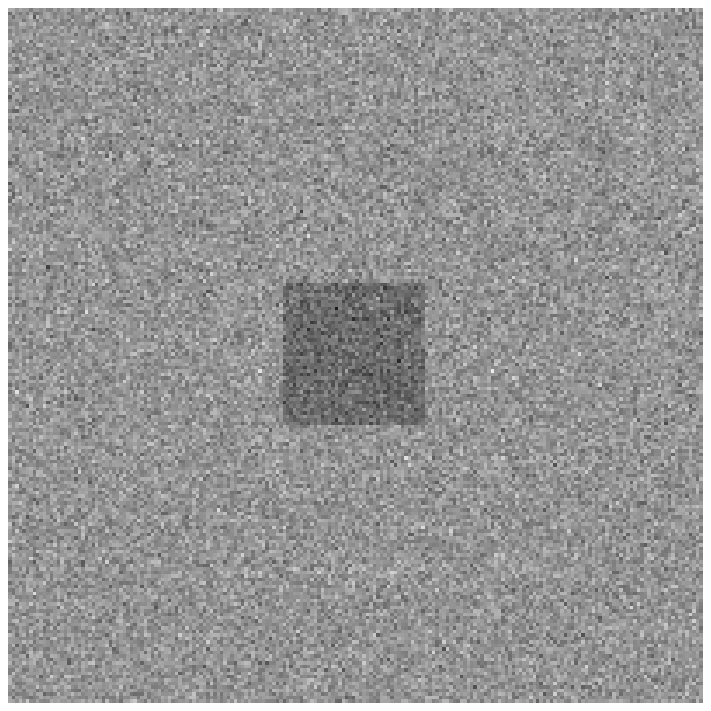}\\%
		\footnotesize Original
	\end{minipage}
	\hfill
	\begin{minipage}[b]{.2\linewidth}
		\centering
		\includegraphics[width=0.8\linewidth]{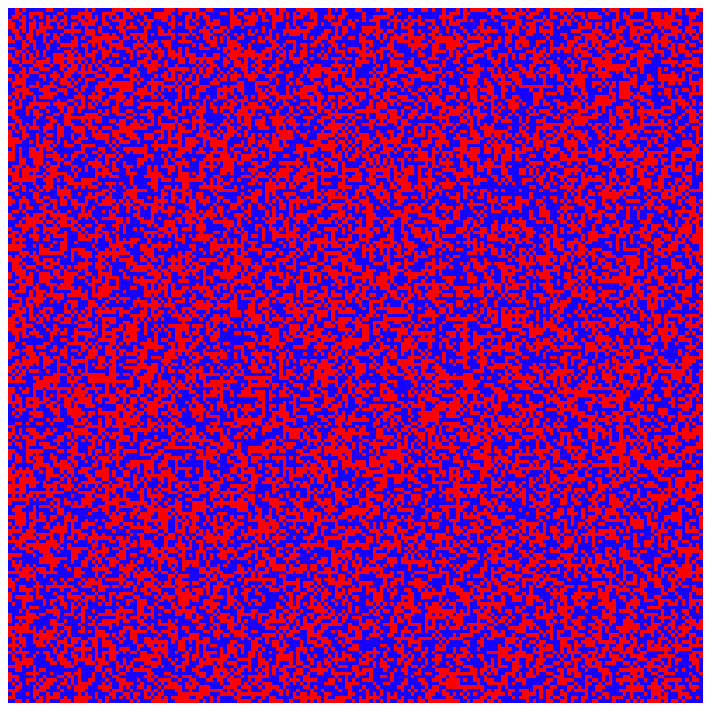}\\%
		\footnotesize Initialization
	\end{minipage}
	\hfill
	\begin{minipage}[b]{.2\linewidth}
		\centering
		\includegraphics[width=0.8\linewidth]{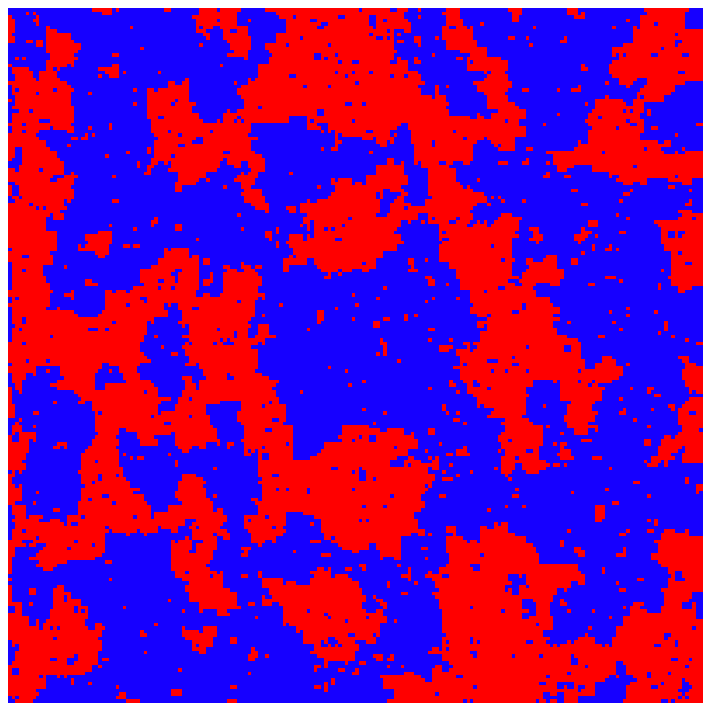}\\%
		\footnotesize Gibbs
	\end{minipage}
	\hfill
	\begin{minipage}[b]{.2\linewidth}
		\centering
		\includegraphics[width=0.8\linewidth]{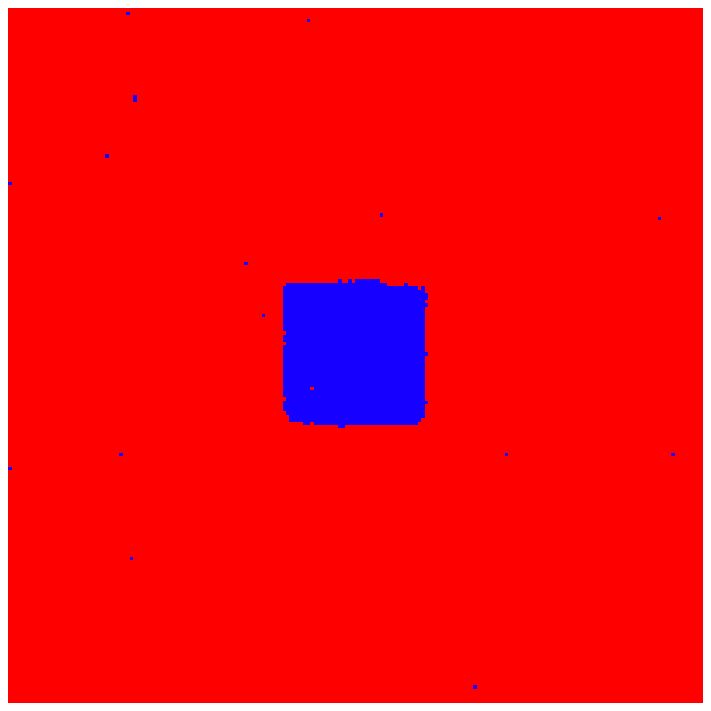}\\%
		\footnotesize GIMH-SS
	\end{minipage}
	\hfill
	\vspace{\vspacecapt}
	\caption{An example of the Gibbs sampling getting stuck in a local extrema as compared to the presented GIMH-SS.}
	\label{fig:gibbs}
\end{figure}

Gibbs sampling is a special case of MH-MCMC, where the acceptance ratio,
$\alpha^{(t)}$, evaluates to one at every iteration. Gibbs proposals typically select a random dimension
and sample from the target distribution conditioned on all other
dimensions. Empirical evidence (as shown in Figure \ref{fig:gibbs})
indicates that when learning appearance models (e.g. in
\cite{Kim2005}), Gibbs sampling exhibits slow convergence times.
Block Gibbs sampling, where a
group of dimensions are sampled conditioned on all
other dimensions, allows for larger moves. The proposed algorithm is related
to this type of Gibbs sampling.

%% file: sampler.tex
\input{sampler-mask.tex}

Sampling a perturbation, $f$, can then be decomposed into sampling a range followed by a value within that range. Most energy functionals only depend on the sign of pixels, which conveniently is determined solely by the range. If we choose $p_{F\vert R}(f \vert r)=\frac{1}{\beta_r}$ (i.e. a uniform distribution), we can write the perturbation likelihood as
\begin{equation}
	p_{FR\vert XM\Phi}(f \vert m,\levelset{t-1})
	 =  \frac{1}{\beta_r} p_{R\vert XM\Phi}(r\vert m,\levelset{t-1}),
	\label{eqn:proposal2}
\end{equation}
where $\beta_r$ is the width or range $r$ (note that $r$ is deterministic conditioned on $f$). Because the value of $f$ within a range $r$ does not affect the sign of the resulting level-set function, the proposed labeling can be expressed as
\begin{equation}
	\hatlabelset{t}(m,r) = \sign{\hatlevelset{t}(m, f\sim p_{F \vert R}(f\vert r)}.
\end{equation}
By choosing the following Gibbs-like proposal for the ranges,
\begin{equation}
	p_{R\vert XM\Phi}(r\vert m,\levelset{t-1}) \propto \beta_r \, \pphi{\hatlabelset{t}(r,m)},
	\label{eqn:proposal-range}
\end{equation}
one can verify via Equations \ref{eqn:hastings}, \ref{eqn:proposal1}, and \ref{eqn:proposal2} that the Hastings ratio evaluates to one at every iteration.

\begin{figure}
	\centering
	\subfloat[]{\label{fig:mask_distributions1}\includegraphics[width=0.3\linewidth]{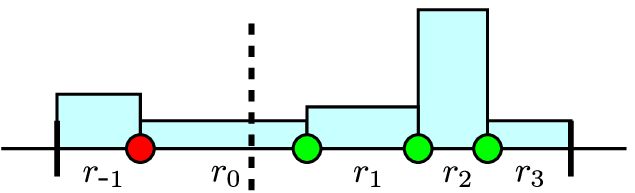}}\hspace*{10pt}\hspace*{20pt}
	\subfloat[]{\label{fig:mask_distributions2}\includegraphics[width=0.3\linewidth]{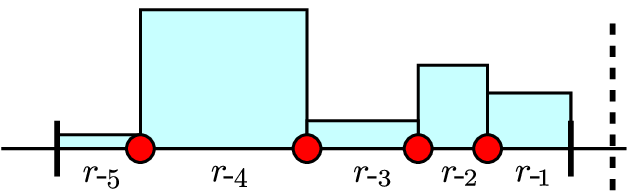}}
	\caption{Example distributions for where the zero-height line can fall. (a) shows an example of a valid proposal, and (b) shows an example of an invalid proposal.}
	\label{fig:mask_distributions}
\end{figure}

One subtlety exists in the selection of endpoints for the first and last range. Since these two ranges extend to $\pm\infty$, their corresponding $\beta$ is also $\infty$. We therefore restrict both of these ranges to be of some finite length $\beta_\infty$. While any finite value will suffice, in practice, we choose $\beta_\infty$ to be $1$. With the endpoint selection for $\beta_\infty$, another complication arises. Consider the illustrative distributions for the zero-height line shown in Figure \ref{fig:mask_distributions}.
In these examples, some constant $\beta_\infty$ is chosen, and the current zero-height line is marked with the dotted line. We can consider these two distributions as two different mask selections. Figure \ref{fig:mask_distributions1} shows what we call a \textit{valid} mask, in which the current zero-height line falls within the non-zero probability space. Figure \ref{fig:mask_distributions2}, however, shows what we call an \textit{invalid} mask, in which the current zero-height line falls outside the non-zero probability space. Here, one could propose a forward perturbation such that the resulting proposed zero-height line falls in the non-zero probability space. However, the backward transition to bring the proposed zero-height line back to the current zero-height line would be zero because it lies in the zero-probability space. We note that if $\beta_\infty$ is chosen to be some constant $C$, and all pixels are initialized to have height $\sizeof{\varphi_i} < C$, then every possible (initial) mask will be valid. As the chain progresses, some masks will become invalid, in which case we only sample valid masks. Because any mask that contains both positive and negative pixels is valid, there are always valid masks to sample from.

\subsection{Algorithm Description}
The resulting algorithm, called the Gibbs-Inspired Metropolis Hastings Shape Sampler (GIMH-SS), is described by the following steps: (1) Sample a circular mask, $m$, with a random center and radius; (2) Sample a range, $r$, according to Equation \ref{eqn:proposal-range}; (3) Sample a perturbation, $f$, uniformly in this range; (4) Compute $\hatlevelset{t}(m,f)$ using Equation \ref{eqn:proposal-form}; (5) Repeat from Step 1 until convergence.

When calculating the likelihood of each range in Step 2, only a value proportional to the true likelihood is needed. Normalization can be performed after the fact because all ranges are enumerated. Equivalently, the energy, which is related to the likelihood via a logarithm, only needs to be calculated up to a constant offset. We therefore compute the change in the energy by including or excluding the pixels in each range. This can be efficiently computed for all ranges using two passes. On each pass, we begin at the range $r_0$, setting this energy to zero and choosing to either go in the positive (first pass) or negative (second pass) direction. We then consider $r_{\pm 1}$ and compute the change in energy when the the one pixel changes sign. In graph-based energies, if an $n$-affinity graph is used, only $n$ computations are required at each range evaluation, one for each edge. This procedure is iterated until all ranges are computed (i.e. up through $r_{\pm n^{\pm}}$).

Unlike graph-based energy functionals, the curve length penalty used in PDE-based energy functionals can be a non-local and fairly expensive computation. In optimization-based segmentation algorithms, the evolution of the level-set is based on the gradient of the curve length (i.e. the curvature) which can efficiently be calculated locally. If a subpixel-accurate curve length calculation is required, then the energy functional will indeed depend on the value of the perturbation, $f$, within a particular range, $r$. As an approximation, we compute a pixel-accurate curve length by first setting all pixels on the boundary to have height $\pm 0.5$, initializing a signed-distance function, and then calculating the curve length via
\begin{equation}
	\oint_{\mathcal{C}}dl
	= \int_\Omega \delta(\varphi_i) \sizeof{\nabla \varphi_i} di
	\approx \sum_{i} \hat\delta(\varphi_i) \sizeof{\nabla\varphi_i},
	\label{eqn:clp}
\end{equation}
and the following discrete approximation to the $\delta(\cdot)$ function
\begin{equation}
	\hat\delta(\varphi_i) =
	\begin{cases}
		1 & \left\vert\varphi_i\right\vert \leq 1\\
		0 & \text{else}
	\end{cases}
\end{equation}

\begin{figure}
	\hfill%
	\subfloat[]{\label{fig:clp_1}\includegraphics[width=0.125\linewidth]{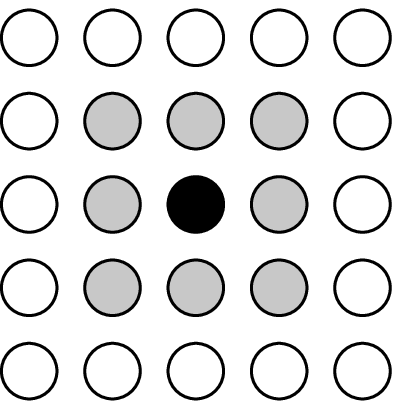}}\hfill%
	\subfloat[]{\label{fig:clp_2}\includegraphics[width=0.125\linewidth]{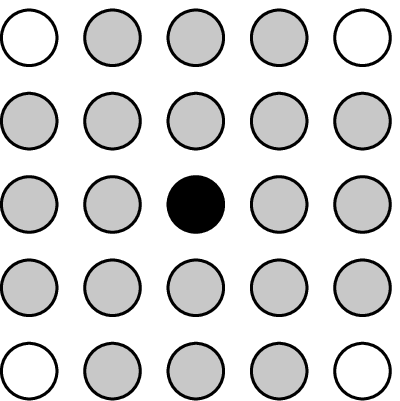}}\hfill%
	\vspace{\vspacecapt}
	\caption{(a) Signed distance function dependence and (b) curve length penalty dependence.}
\end{figure}

If a first-order accurate signed-distance function is obtained using a fast marching method, the height at a particular pixel depends only on pixels in a 3x3 neighborhood. This dependence relationship is illustrated in Figure \ref{fig:clp_1}, where changing the sign of the center black pixel will affect the height of the signed distance function at all the gray pixels. From equation \ref{eqn:clp}, the local curve length calculation depends on the magnitude of the gradient of the level-set function. If a centered finite-difference is used for the $x$ and $y$ directions, then the local curve length computation depends on neighbors above, below, left, and right. Thus, if the heights are changed for the gray pixels in Figure \ref{fig:clp_1}, the curve length computation changes for all the gray pixels in Figure \ref{fig:clp_2}. The change in the curve length for changing the sign of the center pixel is consequently a function of the sign of the 21 pixels in Figure \ref{fig:clp_2}. We precompute all possible $2^{21}$ ($~2$ million) combinations so that the curve length penalty can be efficiently computed by a simple table lookup.

\subsection{Convergence Speeds}
\begin{figure}
	\hfill
	\begin{minipage}[b]{.2\linewidth}
		\centering
		\includegraphics[width=\linewidth]{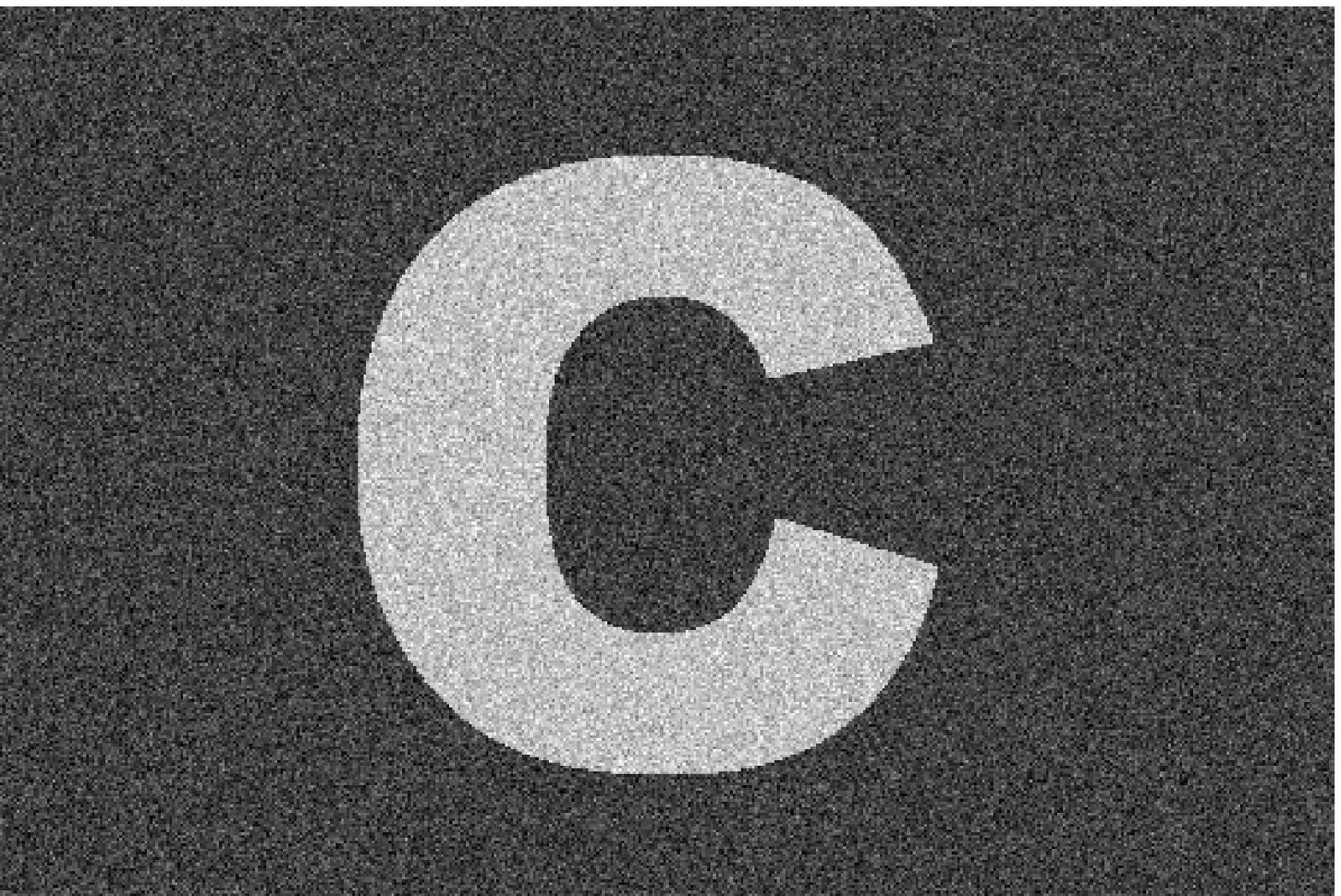}\\\vspace{15pt}%
	\end{minipage}
	\begin{minipage}[b]{.3\linewidth}
		\centering
		\includegraphics[width=\linewidth]{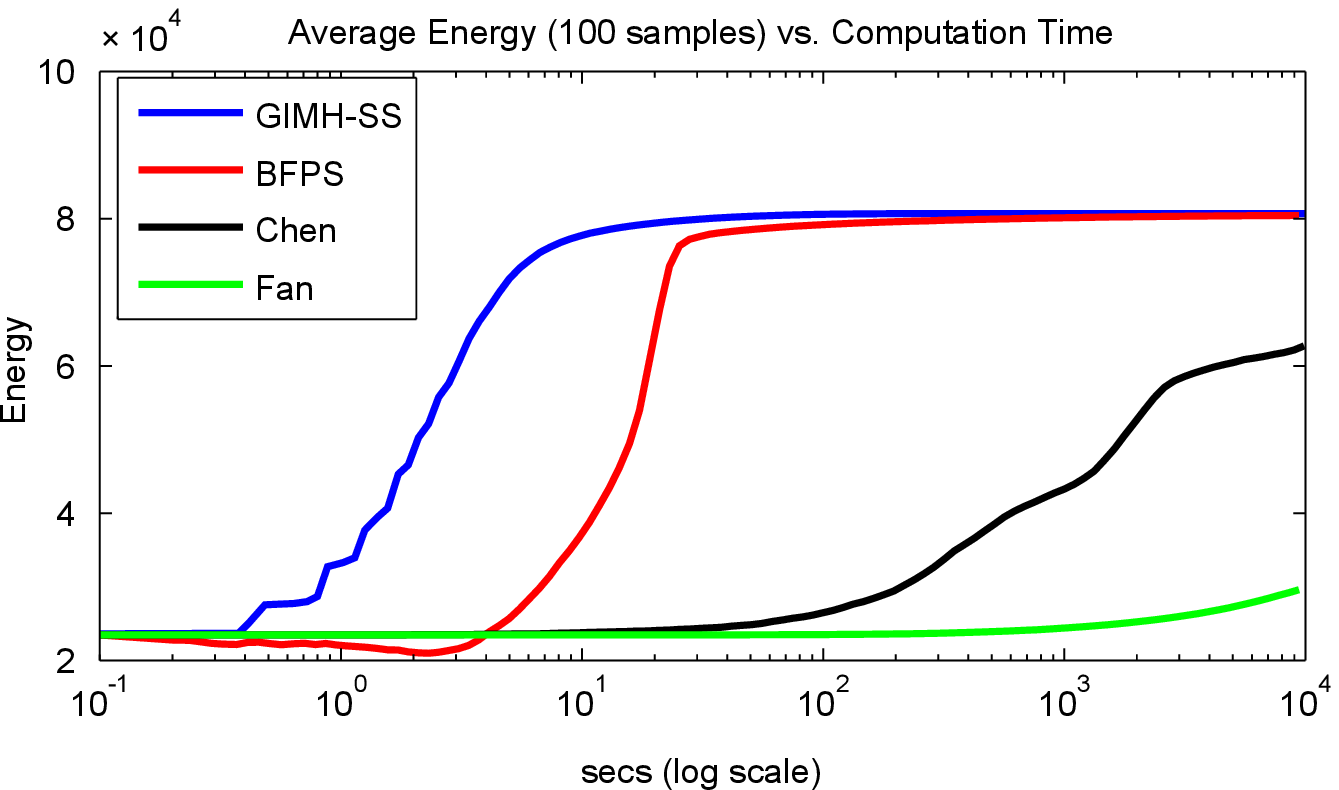}%
	\end{minipage}
	\hfill%
	\vspace{3\vspacecapt}
	\caption{Average energy across 100 sample paths for various shape sampling algorithms. Original image shown on the left.}
	\label{fig:comptime}
\end{figure}

To illustrate the performance of GIMH-SS, we run the sampling algorithm on the image shown in Figure \ref{fig:comptime}. We note that this is the same image used in the comparison of \cite{Chang2011}. We run each of the four algorithms for $10^4$ seconds. As in \cite{Chang2011}, an approximate algorithm is used for \cite{Fan2007} and \cite{Chen2009} that does not necessarily ensure detailed balance. Certain random initializations require topological changes to converge to the correct segmentation, which neither \cite{Fan2007} and \cite{Chen2009} can not accomodate. In these situations, samples are accepted with a certain probably instead of instantly being rejected. We note that without this approximation, the algorithms would \textbf{never} converge to the correct solution. While BFPS exhibits running times that are orders of magnitude faster than \cite{Fan2007} and \cite{Chen2009}, GIMH-SS is almost an order of magnitude faster than BFPS. The average energy obtained using GIMH-SS both rises faster and converges faster than BFPS. Additionally, the gradient of the energy functional (which potentially can be difficult to evaluate) is never needed in the proposal.
These results combined with those found in \cite{Chang2011} indicate that GIMH-SS draws samples approximately $10$x faster than BFPS, $10^5$x faster than \cite{Chen2009}, and $10^6$x faster than \cite{Fan2007}.

\subsection{Relation to Block Gibbs Sampling}
Block Gibbs sampling first selects a mask of pixels (or dimensions) to changes, and then samples from the target distribution conditioned on all other pixels. In binary segmentation, a block of size $\sizeof{m}$ would require one to evaluate an exponential number ($2^{\sizeof{m}}$) of different configurations, which is intractable for a reasonably sized mask.

The GIMH-SS algorithm similarly selects a mask of pixels. The level-set function orders the masked pixels, in that, if a pixel of height $+h$ changes sign, all pixels of height $0<\varphi_i<h$ must as well. Consequently, this algorithm samples from a subset of the conditional distribution, resulting in a linear number ($\sizeof{m}+1$) of different configurations. We note that ergodicity is ensured because the ordering of pixels by the level-set function changes over time.

%% file: sampler-mask.tex
\begin{figure}
	\centering
	\includegraphics[width=0.5\linewidth]{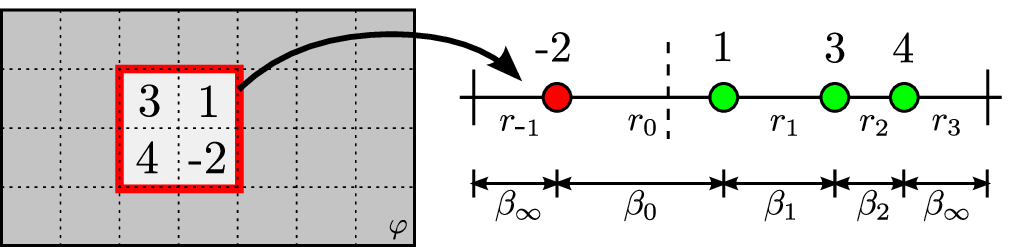}
	\vspace{\vspacecapt}
	\caption{Mapping the masked level-set function (left) onto the real line (right). Possible ranges, $r$, and widths, $\beta$, are shown on the right.}
	\label{fig:mask}
\end{figure}

BFPS \cite{Chang2011} achieves a high acceptance ratio at each iteration by biasing the proposal by the gradient of the energy functional. Here, we design a similar proposal such that samples are accepted with certainty. Given some previous level-set function, $\levelset{t-1}$, we generate a proposal, $\hatlevelset{t}$ with the following steps: (1) sample a mask, $m$, that selects a subset of pixels and (2) add a constant value, $f$, to all pixels within this mask. We can express this as
\begin{equation}
	\hatlevelset{t}(m,f) = \levelset{t-1} + f \cdot m,
	\label{eqn:proposal-form}
\end{equation}
where the mask, $m$, is a set of indicator variables with the same size as $\varphi$. The support of the mask can be of any shape and size, though in practice we use circles of random center and radius. Deferring the choice of $f$, we write the proposal likelihood as 
\begin{equation}
	q(\hatlevelset{t} \vert \levelset{t-1},x) = p_M(m) \, p_{F\vert XM\Phi }(f \vert m,\levelset{t-1}).
	\label{eqn:proposal1}
\end{equation}
Figure \ref{fig:mask} shows a notional mask overlaid on a level-set function, $\varphi$, with the height of $\varphi$ plotted on the real line for pixels in the support of the mask.
The dotted line marks the zero-height which splits pixels into foreground and background. As all pixels in the support of the mask are summed with the same constant, $f$, choosing an $f$ is equivalent to shifting the zero-height by $-f$. The resulting proposed zero-height can fall into one of $\left(\sizeof{m}+1\right)$ different \textit{ranges}, where $\sizeof{m}$ counts the pixels in the support of the mask. The random range, $R$, of the perturbation can take on integers in $[-n^-, n^+ ]$, where $n^\pm$ count the pixels in the support of the mask that have height $\varphi \gtrless 0$.

%% file: topology.tex
In this section, we extend GIMH-SS to a topology-controlled sampler. The topology of a continuous, compact surface is often described by its genus (i.e. the number of ``handles''). Digital topology \cite{Kong1989} is the discrete counterpart of continuous topology, where regions are represented via binary variables on a grid.

\begin{figure}
	\hfill
	\begin{minipage}[b]{.1\linewidth}
		\centering
		\includegraphics[width=\linewidth]{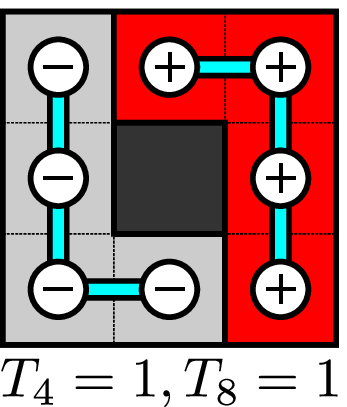}\\%
	\end{minipage}
	\;
	\begin{minipage}[b]{.1\linewidth}
		\centering
		\includegraphics[width=\linewidth]{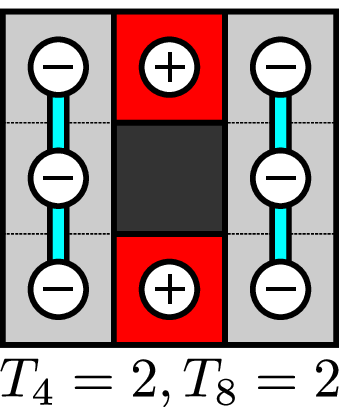}\\%
	\end{minipage}
	\;
	\begin{minipage}[b]{.1\linewidth}
		\centering
		\includegraphics[width=\linewidth]{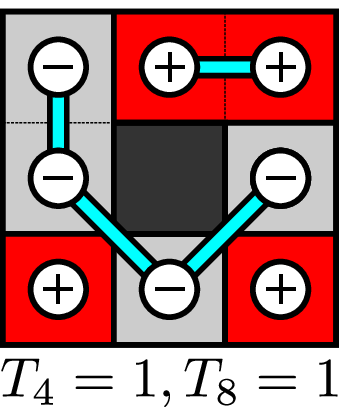}\\%
	\end{minipage}
	\quad
	\begin{minipage}[b]{.2\linewidth}
		\centering
		\includegraphics[width=\linewidth]{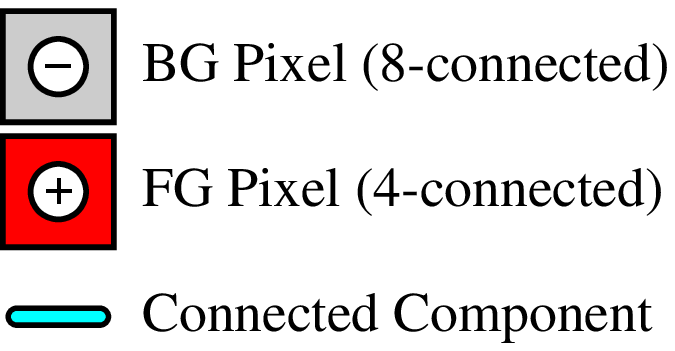}%
		\vspace*{2pt}
	\end{minipage}
	\hfill
	\vspace{\vspacecapt}
	\caption{Examples of topological numbers with $(n, \overline{n})=(4,8)$.}
	\label{fig:topological-numbers}
\end{figure}

In digital topology, \textit{connectiveness} must be defined for the foreground (FG) and background (BG) describing how pixels in a local neighborhood are connected. For example, in 2D, a 4-connected region corresponds to a pixel being connected to its neighbors above, below, left, and right. An 8-connected region corresponds to being connected to the eight pixels in a 3x3 neighborhood. Connectiveness must be jointly defined for the foreground ($n$) and background ($\overline{n}$) to avoid topological paradoxes. As shown in \cite{Kong1989}, valid connectivities for 2D are $(n,\overline{n})\in\left\{(4,8), (8,4)\right\}$.
Given a pair of connectiveness, the topological numbers \cite{Bertrand1994} at a particular pixel, $T_n$ (for the FG) and $T_{\overline{n}}$ (for the BG) count the number of connected components a pixel is connected to in a 3x3 neighborhood. Figure \ref{fig:topological-numbers} shows a few neighborhoods with their corresponding topological numbers.
While these topological numbers reflect topological changes, they do not distinguish splitting or merging of regions from the creation or destruction of a handle. Segonne \cite{Segonne2008} defines two additional topological numbers, $T_n^+$ and $T_{\overline{n}}^+$, which count the number of \textit{unique} connected components a pixel is connected to over the entire image domain. $T_n^+$ and $T_{\overline{n}}^+$ depend on how pixels are connected outside of the 3x3 region and allow one to distinguish all topological changes.

By labeling each connected component in the foreground and background, Segonne shows that $T_n^+$ can be computed efficiently when a pixel is added to the foreground and $T_{\overline{n}}^+$ when a pixel is added to the background. In 3D, one \textit{must} calculate $T_{\overline{n}}^+$ when adding a pixel to the foreground, which can be computationally expensive. We show here that this calculation is not necessary in 2D.

\begin{figure}
	\centering
	\includegraphics[width=0.6\linewidth]{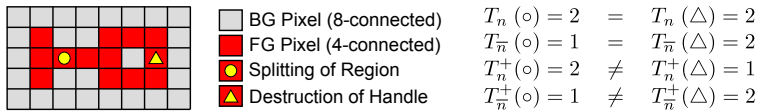}
	\vspace{\vspacecapt}
	\caption{Splitting a region vs. destroying a handle while a pixel is added to BG. Topologal numbers are shown on right.}
	\label{fig:handle-id}
\end{figure}

Consider the two pixels marked by $\circ$ and $\triangle$ in Figure \ref{fig:handle-id}.
Removing the $\circ$ pixel from the foreground splits the region and removing the $\triangle$ pixel destroys a handle. We emphasize that in this 2D case, both $T_n^+(\circ)\neq T_n^+(\triangle)$ \textit{and} $T_{\overline{n}}^+(\circ)\neq T_{\overline{n}}^+(\triangle)$.
In 3D, this is generally not the case because for the $\triangle$ pixel, the two background regions can actually be connected in another 2D slice of the volume.
In fact, in 2D, the destruction of a handle in the foreground corresponds directly to a merging of regions in the background. Likewise, the splitting of regions in the foreground corresponds directly to a creation of a handle in the background. Because of this one-to-one mapping, we do not need to compute the expensive $T_{\overline{n}}^+$ when adding a pixel to the foreground. Table \ref{tbl:topologies} summarizes the topological changes of the foreground and background in 2D as a function of the four topological numbers.
\begin{table}
	\centering
	\begin{tabular}{c c c c || c c | c c}
		& & & & \multicolumn{2}{|c|}{Add to FG} & \multicolumn{2}{|c}{Add to BG}\\
		$T_n$ & $T_n^+$ & $T_{\overline{n}}$ & $T_{\overline{n}}^+$ & FG & BG & FG & BG\\\hline\hline
		0 & 0 & 1 & 1 & CR & CH & DR & DH\\
		1 & 1 & 0 & 0 & DH & DR & CH & CR\\
		1 & 1 & 1 & 1 & - & - & - & -\\
		$\geq 2$ & $<T_n$ & X & X & CH & SR & X & X \\
		$\geq 2$ & $\geq 2$ & X & X & MR & DH & X & X \\
		X & X & $\geq 2$ & $<T_{\overline{n}}$ & X & X & SR & CH \\
		X & X & $\geq 2$ & $\geq 2$ & X & X & DH & MR
	\end{tabular}
	\vspace{\vspacecapt}
	\caption{Topological changes as a function of topological numbers. `C' - Create; `D' - Destroy; `S' - Split; `M' - Merge; `H' - Handle(s); `R' - Region(s); `X' - any value; `-' - no topological change. Omitted configurations are impossible in 2D.}
	\label{tbl:topologies}
\end{table}

\subsection{Topology-Controlled GIMH-SS} In this section, we summarize how to sample from the space of segmentations while enforcing topology constraints (\textit{c.f.} \cite{Chang2012tech} for details). While the goal is similar to \cite{Segonne2008}, a simple alteration of the level-set velocity in an MCMC framework will not preserve detailed balance. A na\"{i}ve approach generates proposals using GIMH-SS and rejects samples that violate topology constraints. Such an approach wastes significant computation generating samples that are rejected due to their topology. We take a different approach: only generate proposed samples from the set of allowable topological changes. Recalling the discussion of ranges in GIMH-SS, one can determine which ranges correspond to allowable topologies and which do not. Ranges corresponding to restricted topologies have their likelihood set to zero in a topology-controlled version of GIMH-SS (TC-GIMH-SS). The methodology presented here treats the topology as a hard constraint. A distribution over topologies could be implemented by weighting ranges based on topologies rather than eliminating restricted ones.

Similar to GIMH-SS, TC-GIMH-SS also makes two passes to calculate the likelihoods of the ranges. Starting at $r_0$, we either proceed in the positive or negative direction. Because $r_0$ corresponds to the range that does not change the sign of any pixels, $r_0$ will never correspond to a topological change. As the algorithm iterates through the possible ranges, a list of pixels that violate a topological constraint is maintained. If this violated list is empty after a range is considered, then the range corresponds to an allowable topology. At each iteration, while any pixel in the violated list is allowed to change sign, it is removed and all neighboring pixels in the violated list are checked again as their topological constraint may have changed. In this process, for each range, one must maintain the labels of each connected component.

%% file: results.tex
\begin{figure}
	\hfill
	\begin{minipage}[b]{.15\linewidth}
		\centering
		\includegraphics[width=0.9\linewidth]{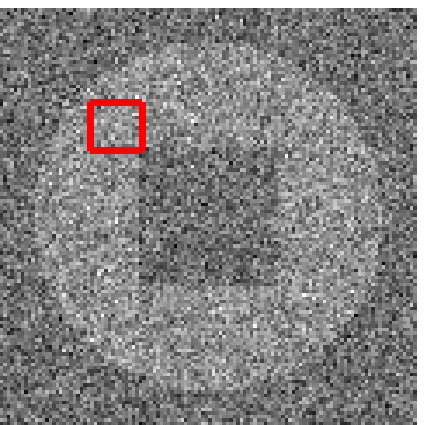}\\%
		\footnotesize Initialization
	\end{minipage}
	\begin{minipage}[b]{.15\linewidth}
		\centering
		\includegraphics[width=0.9\linewidth]{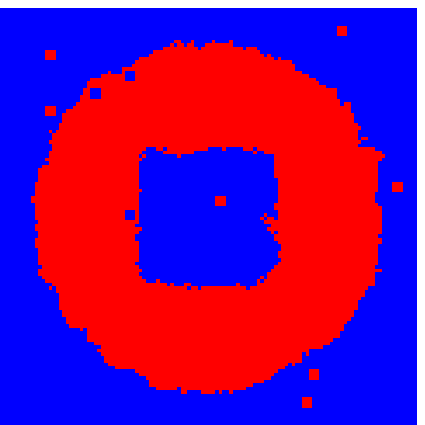}\\%
		\footnotesize GIMH-SS
	\end{minipage}
	\begin{minipage}[b]{.15\linewidth}
		\centering
		\includegraphics[width=0.9\linewidth]{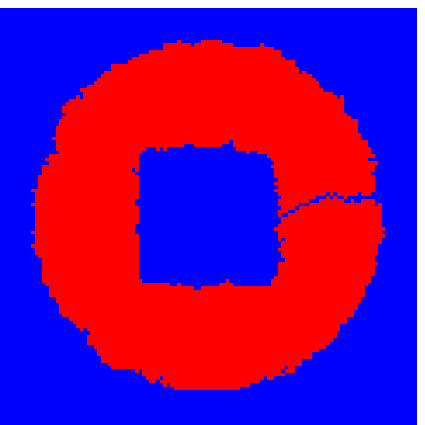}\\%
		\footnotesize TP
	\end{minipage}
	\begin{minipage}[b]{.15\linewidth}
		\centering
		\includegraphics[width=0.9\linewidth]{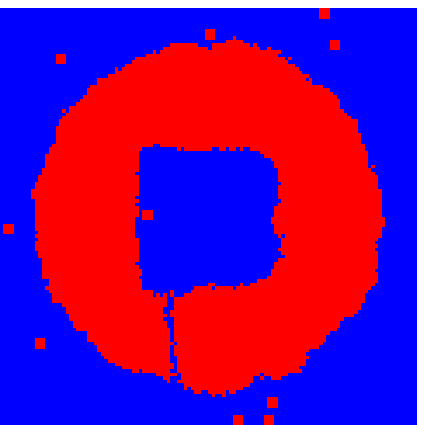}\\%
		\footnotesize GP
	\end{minipage}
	\begin{minipage}[b]{.15\linewidth}
		\centering
		\includegraphics[width=0.9\linewidth]{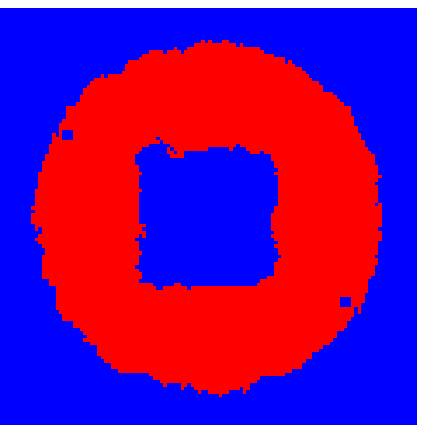}\\%
		\footnotesize CCP
	\end{minipage}
	\hfill
	\vspace{\vspacecapt}
	\caption{Example samples imposing different topology constraints.}
	\label{fig:topologies}
\end{figure}

While TC-GIMH-SS applies to almost any PDE-based or graph-based energy
functional, we make a specific choice for demonstration
purposes. Similar to \cite{Kim2005}, we learn nonparametric
probability densities over intensities and combine mutual information
with a curve length penalty. We impose four different topology
constraints on the foreground: unconstrained (GIMH-SS),
topology-preserving (TP), genus-preserving (GP), and
connected-component-preserving (CCP). The TP sampler does not allow
any topology changes, the GP sampler only allows the splitting and
merging of regions, and the CCP sampler only allows the creation and
destruction of handles. Typical samples from each of these constraints 
are shown in Figure \ref{fig:topologies}. When the topology
constraint is incorrect for the object (e.g. using TP or GP), the
resulting sample may be undesirable (e.g. creating an isthmus
connecting the two connected components of the background). When the
topology is correct, however, more robust results can be obtained. For
example, the CCP constraint removes some noise in the background.

\begin{figure}
\hfill
	\begin{minipage}[b]{.25\linewidth}
		\centering
		\includegraphics[width=0.98\linewidth]{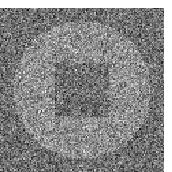}\\
		\scriptsize Original
	\end{minipage}
	\hfill
	\begin{minipage}[b]{.1\linewidth}
		\centering
		FG Init\\
		\vspace*{38pt}
		Random Init\\
		\vspace*{30pt}
	\end{minipage}
	\hfill
	\begin{minipage}[b]{.12\linewidth}
		\centering
		\includegraphics[width=\linewidth]{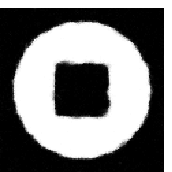}\\\vspace*{2pt}%
		\includegraphics[width=\linewidth]{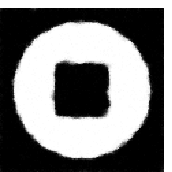}\\%
		\scriptsize GIMH-SS
	\end{minipage}
	\begin{minipage}[b]{.12\linewidth}
		\centering
		\includegraphics[width=\linewidth]{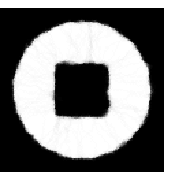}\\\vspace*{2pt}%
		\includegraphics[width=\linewidth]{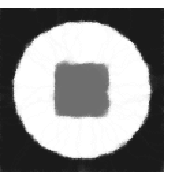}\\%
		\scriptsize TP
	\end{minipage}
	\begin{minipage}[b]{.12\linewidth}
		\centering
		\includegraphics[width=\linewidth]{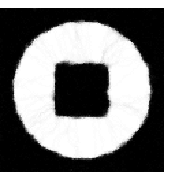}\\\vspace*{2pt}%
		\includegraphics[width=\linewidth]{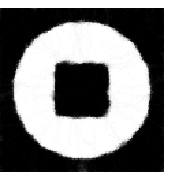}\\%
		\scriptsize GP
	\end{minipage}
	\begin{minipage}[b]{.12\linewidth}
		\centering
		\includegraphics[width=\linewidth]{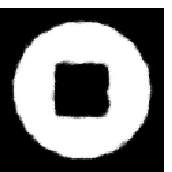}\\\vspace*{2pt}%
		\includegraphics[width=\linewidth]{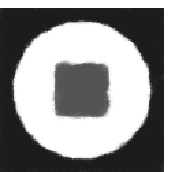}\\%
		\scriptsize CCP
	\end{minipage}
	\hfill
	\vspace{\vspacecapt}
	\caption{Histogram images using different initializations.}
	\label{fig:histograms}
\end{figure}

The usefulness of the topology constraint relies on a valid
initialization. Histogram images \cite{Fan2007} display the marginal
probability that a pixel belongs to the foreground or background where
lighter colors equate to higher foreground likelihood. The computed
histogram images are shown in Figure \ref{fig:histograms} for each of
the topology constraints.
We initialize the samples either using a random circle containing the
foreground (FG Init), or a random circle placed anywhere in the image
(Random Init).  While not always true, incorrect topology constraints are sometimes mitigated
when looking at marginal statistics. For example, in the FG
initialization case, the isthmus in the TP and GP constraints is no
longer visible. Additionally, if
the initialization only captures one connected component of the
background (which is possible with random initialization), certain
samplers prohibiting splitting regions (TP and CCP) will not be able
to capture the entire region. This is reflected in the histogram image
with the gray center.

\begin{figure}
	\hfill
	\begin{minipage}[b]{.12\linewidth}
		\centering
		\includegraphics[width=\linewidth]{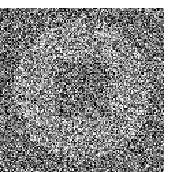}\\\vspace*{4pt}
		\includegraphics[width=\linewidth]{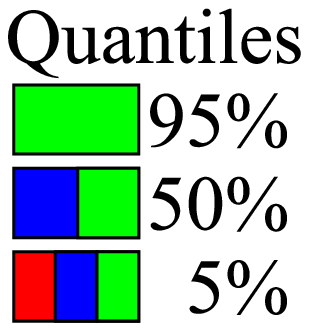}\\\vspace*{5pt}
	\end{minipage}
	\hfill
	\begin{minipage}[b]{.12\linewidth}
		\centering
		\includegraphics[width=\linewidth]{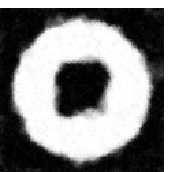}\\\vspace*{2pt}%
		\includegraphics[width=\linewidth]{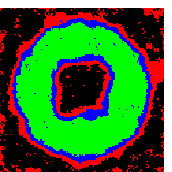}\\
		\footnotesize GIMH-SS
	\end{minipage}
	\begin{minipage}[b]{.12\linewidth}
		\centering
		\includegraphics[width=\linewidth]{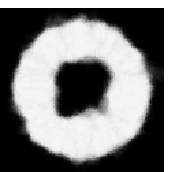}\\\vspace*{2pt}%
		\includegraphics[width=\linewidth]{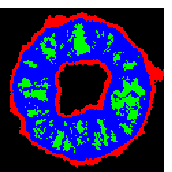}\\
		\footnotesize TP
	\end{minipage}
	\begin{minipage}[b]{.12\linewidth}
		\centering
		\includegraphics[width=\linewidth]{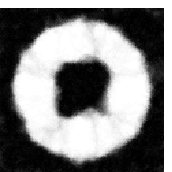}\\\vspace*{2pt}%
		\includegraphics[width=\linewidth]{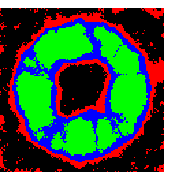}\\
		\footnotesize GP
	\end{minipage}
	\begin{minipage}[b]{.12\linewidth}
		\centering
		\includegraphics[width=\linewidth]{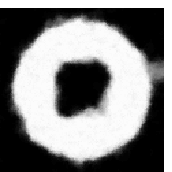}\\\vspace*{2pt}%
		\includegraphics[width=\linewidth]{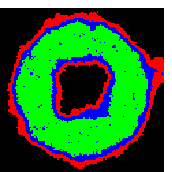}\\
		\footnotesize CCP
	\end{minipage}
	\hfill
	\vspace{\vspacecapt}
	\caption{Low SNR image using FG Init. Histograms are shown above, and quantiles (thresholded histograms) are shown below.}
	\label{fig:lowsnr}
\end{figure}

Consider the low SNR image of Figure \ref{fig:lowsnr}. Chang
\textit{et al} \cite{Chang2011} demonstrated that sampling improves
results when compared to optimization based methods in low SNR
cases. When the problem is less likelihood dominated, as in this case,
the prior has greater impact. The top row of Figure \ref{fig:lowsnr}
shows the histogram images obtained using each of the topological
constraints. One can see remnants of the isthmuses in the TP and GP
cases.
Thresholding the normalized histogram image at a value $t$ reveals the
$t^\text{th}$ quantile of the segmentation. For example, if $t=0.9$,
the resulting foreground region of the thresholded histogram contains
pixels that are in the foreground for at least 90\% of the samples. We
show the 95\%, 50\%, and 5\% quantile segmentations in Figure
\ref{fig:lowsnr}. Since reducing the threshold never shrinks
the foreground segmentation, we can overlay these quantiles on top of
each other. In the 5\% quantile, we can clearly see the isthmuses in
the TP and GP cases. This result is poor because the wrong topology
(i.e. the wrong prior) was used. However, if we use the CCP
constraint, we are able to improve results as compared to the
unconstrained case by removing a lot of the background noise.

\begin{figure}
	\hfill
	\begin{minipage}[b]{.155\linewidth}
		\centering
		\includegraphics[width=\linewidth]{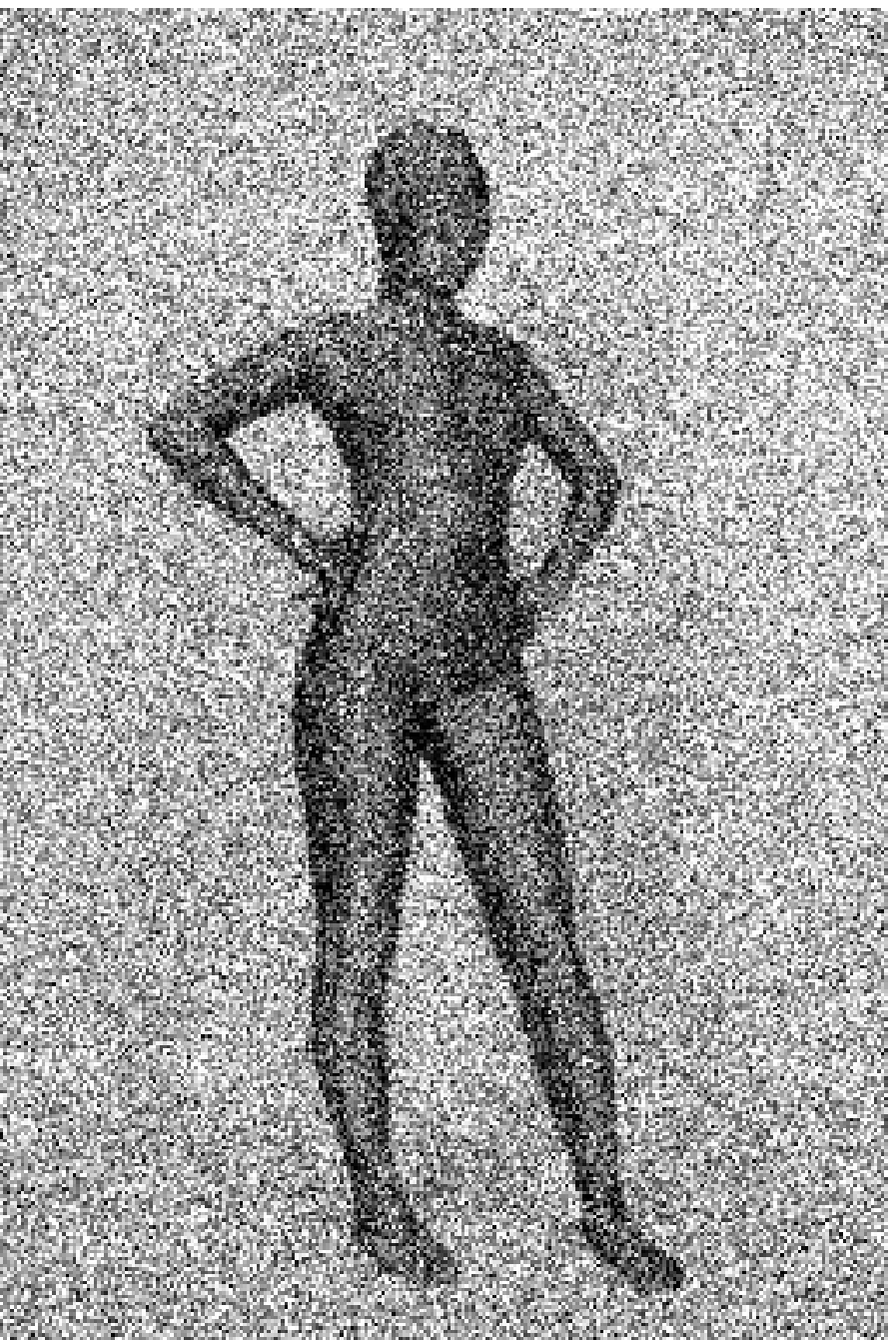}\\%
		\includegraphics[width=\linewidth]{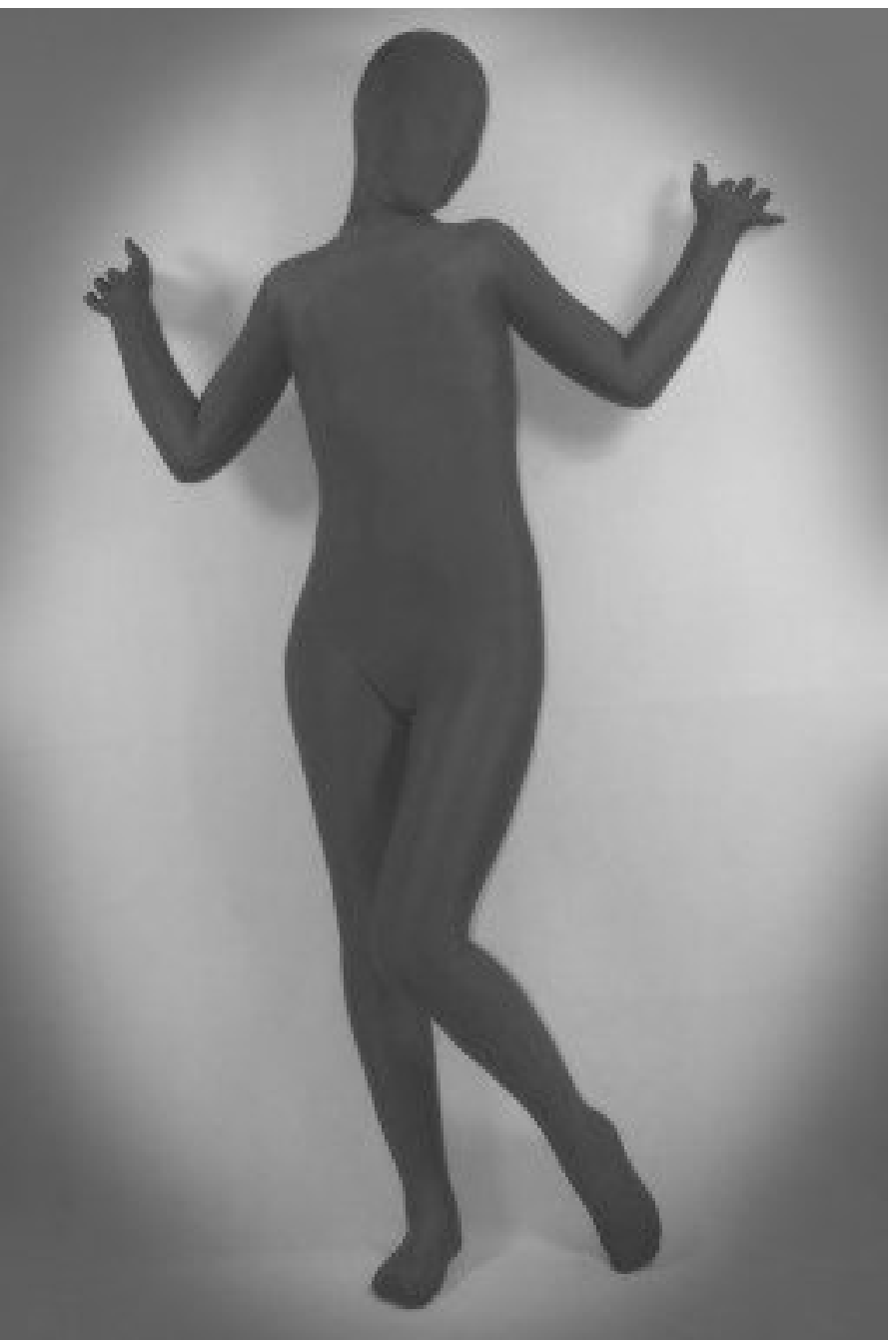}\\
		\footnotesize Original
	\end{minipage}
	\begin{minipage}[b]{.155\linewidth}
		\centering
		\includegraphics[width=\linewidth]{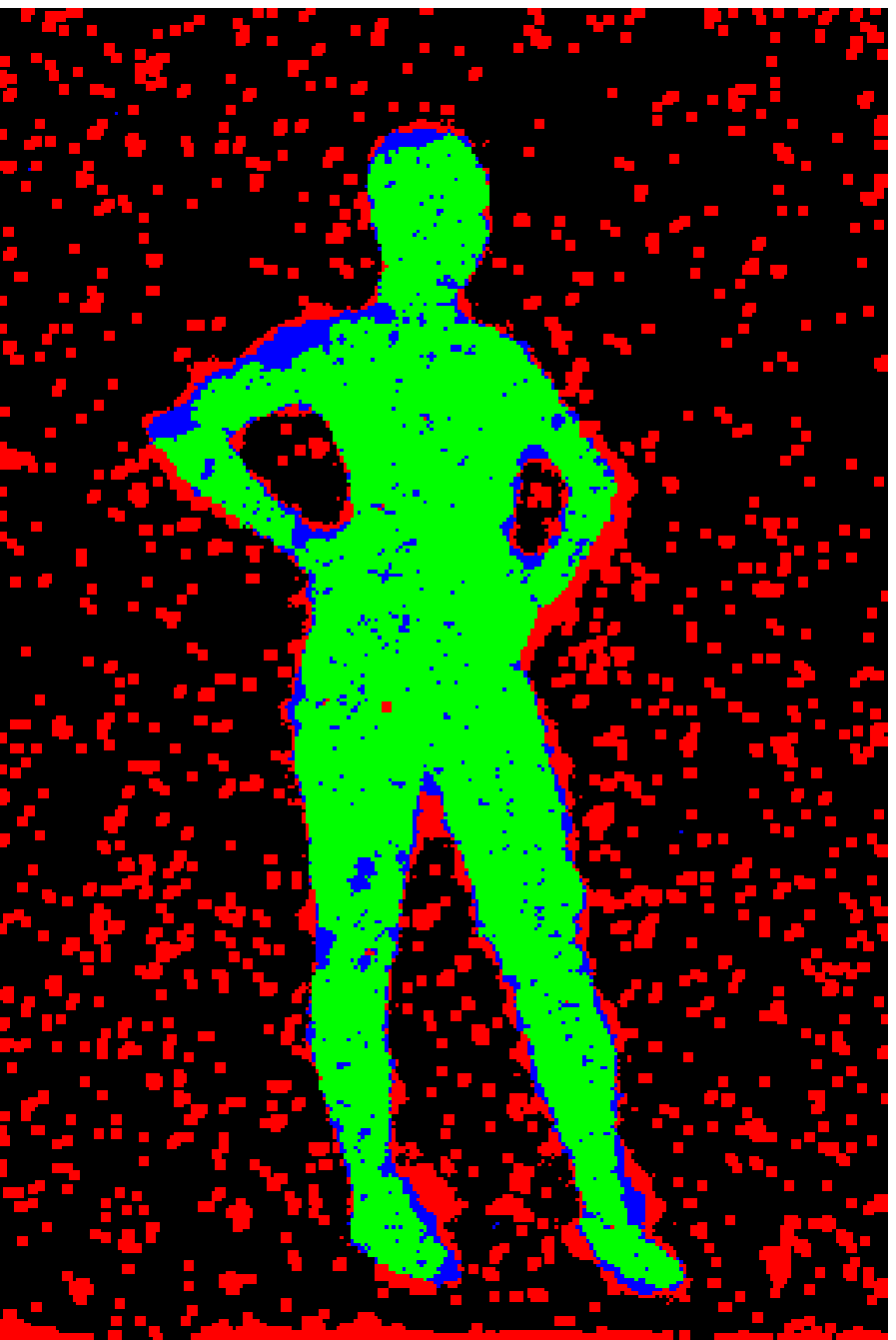}\\%
		\includegraphics[width=\linewidth]{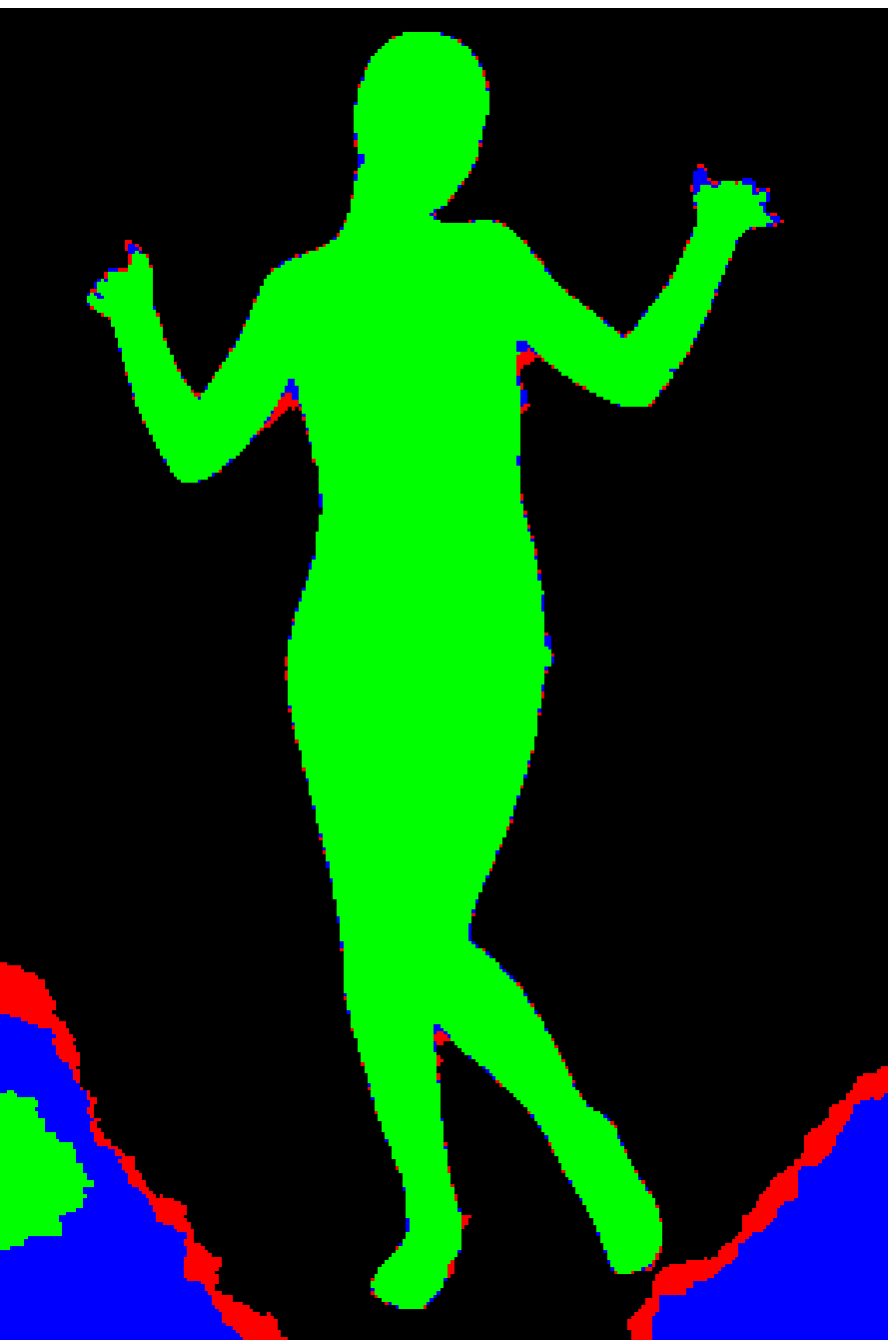}\\
		\footnotesize GIMH-SS
	\end{minipage}
	\begin{minipage}[b]{.155\linewidth}
		\centering
		\includegraphics[width=\linewidth]{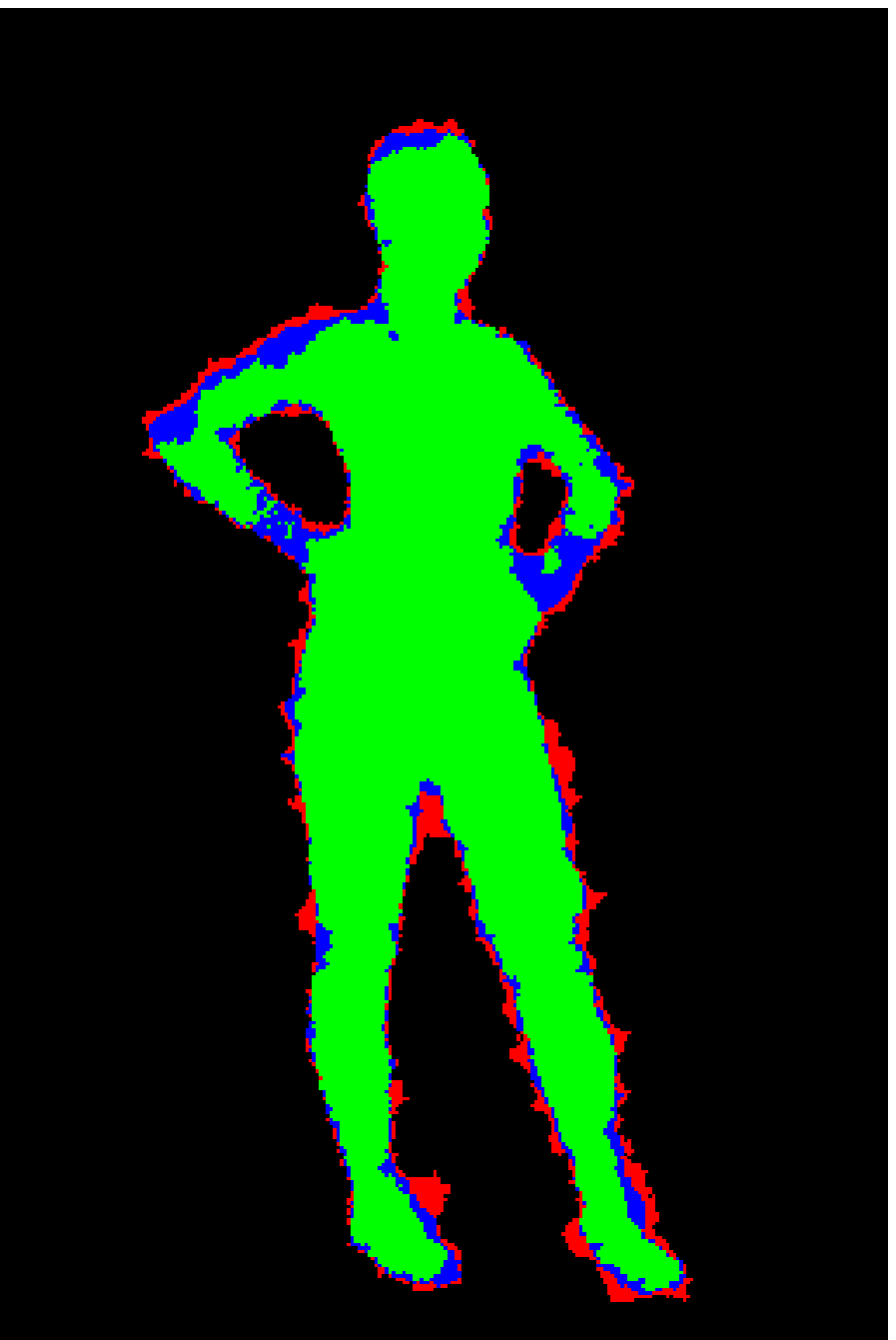}\\%
		\includegraphics[width=\linewidth]{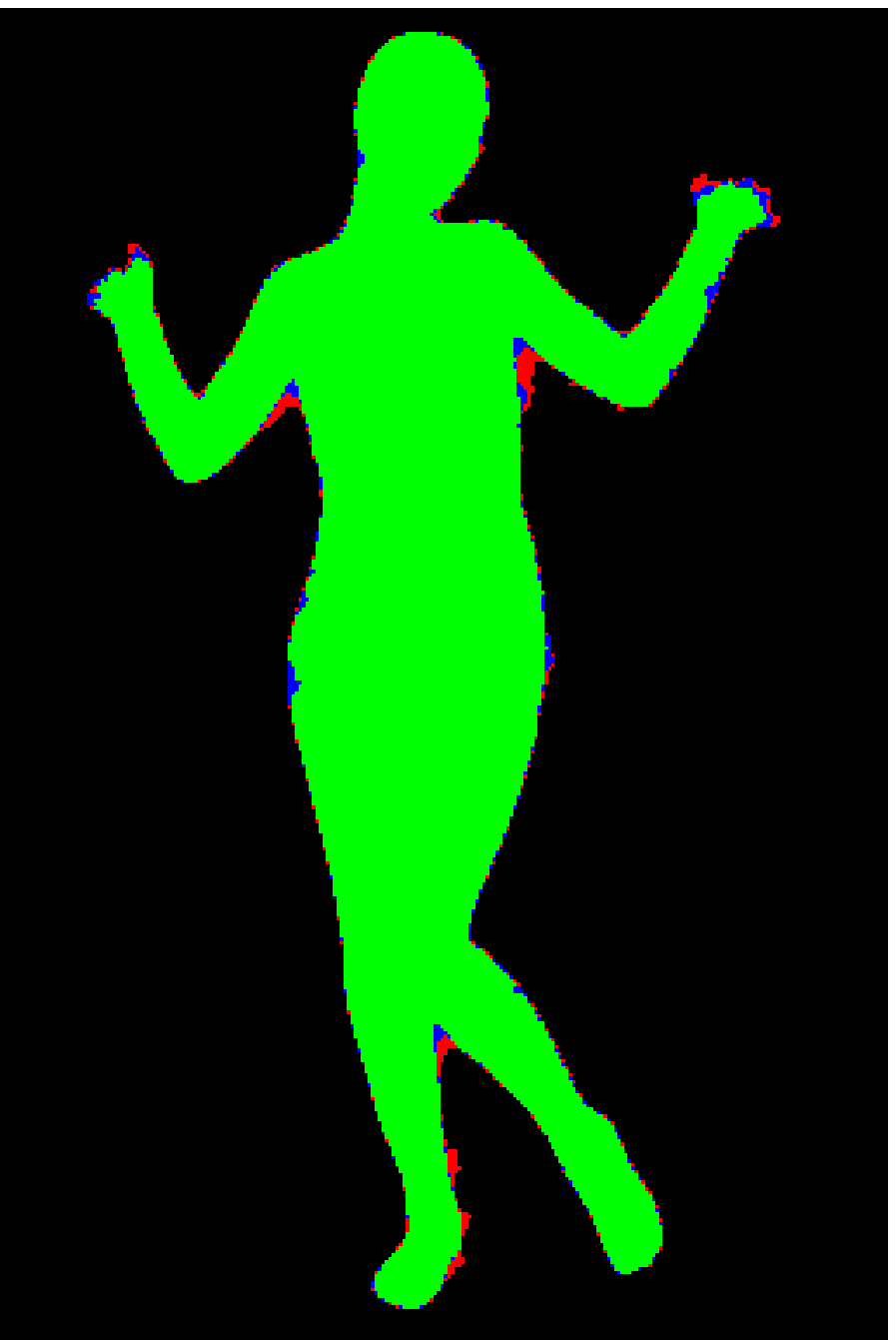}\\
		\footnotesize TP
	\end{minipage}
	\begin{minipage}[b]{.155\linewidth}
		\centering
		\includegraphics[width=\linewidth]{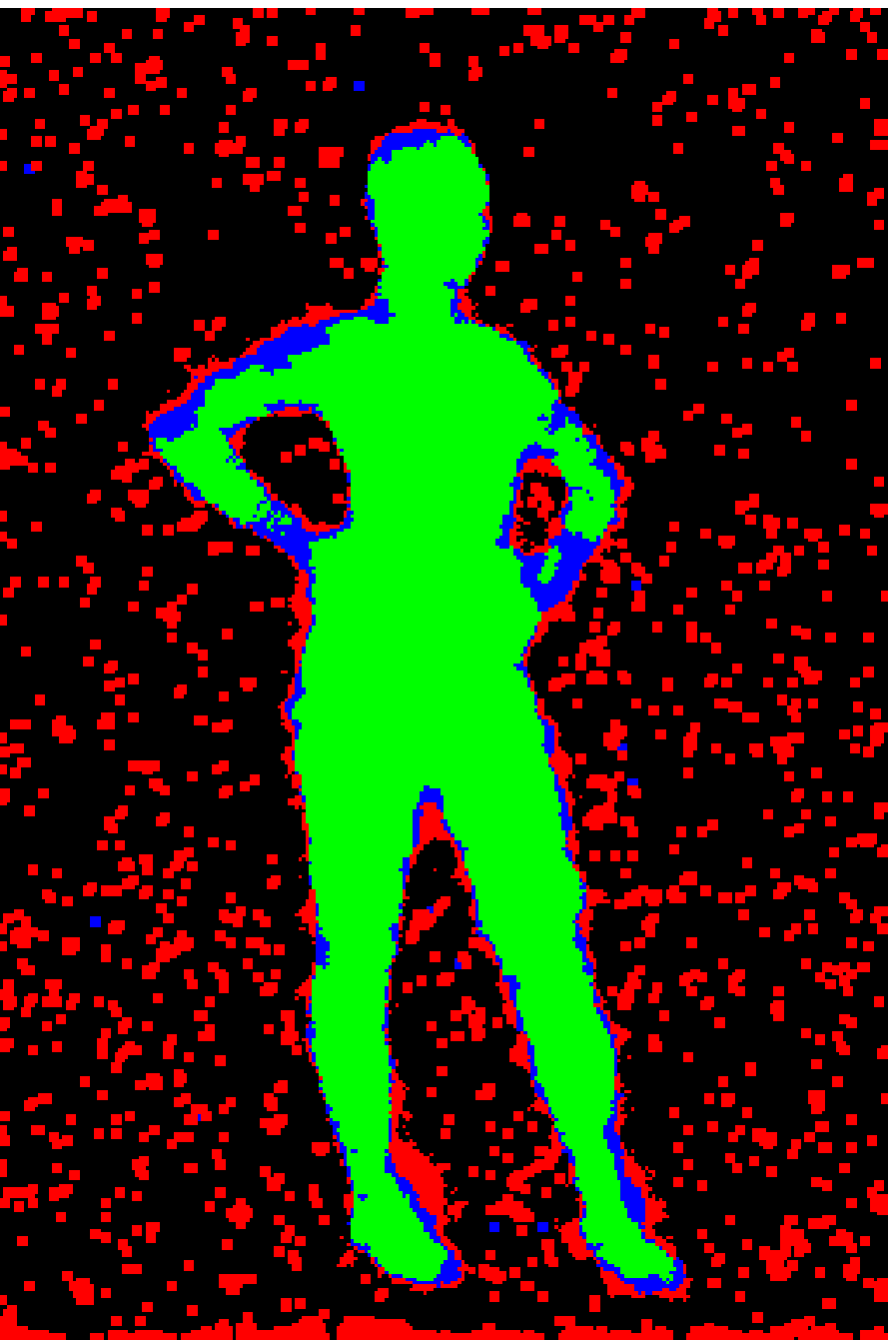}\\%
		\includegraphics[width=\linewidth]{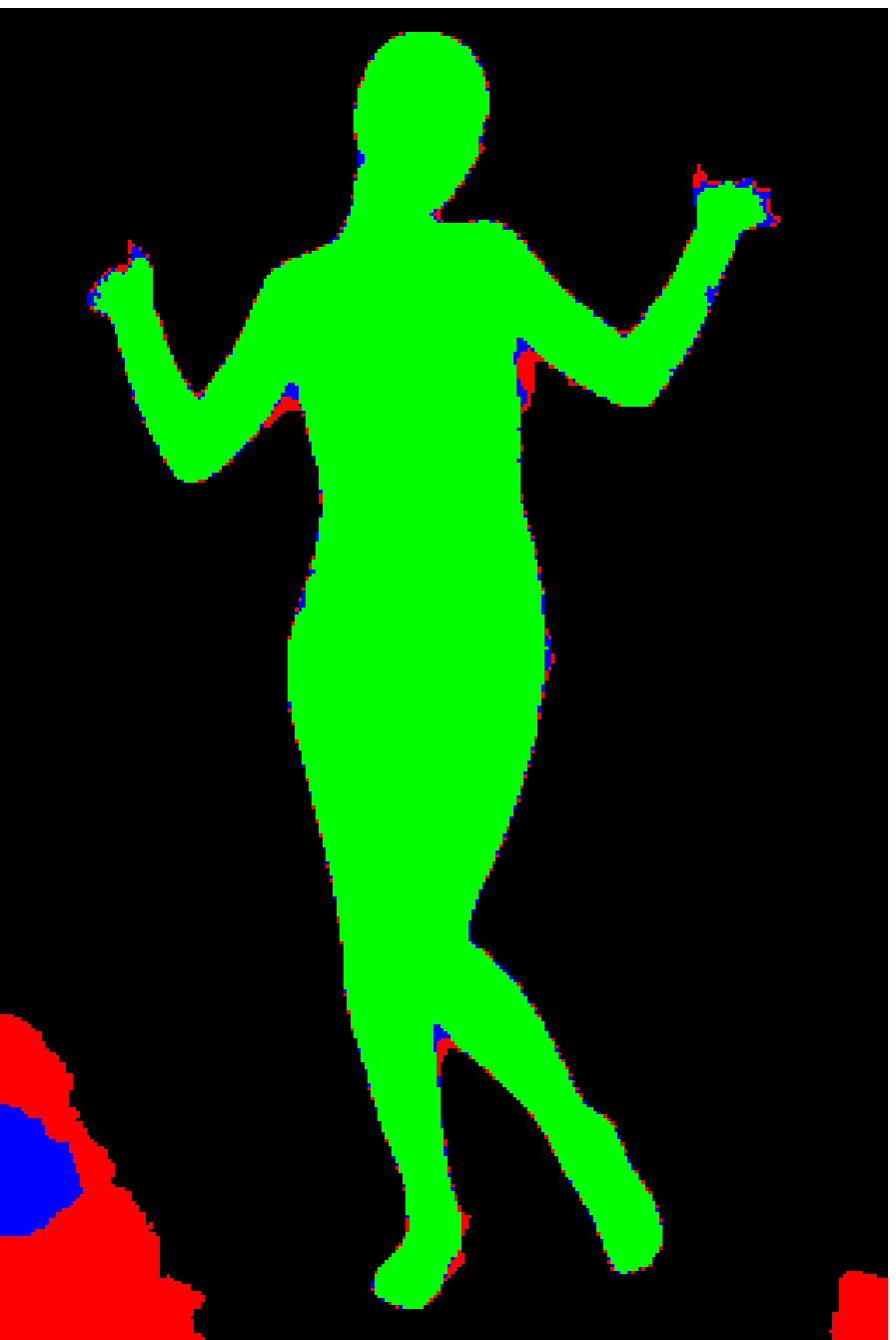}\\
		\footnotesize GP
	\end{minipage}
	\begin{minipage}[b]{.155\linewidth}
		\centering
		\includegraphics[width=\linewidth]{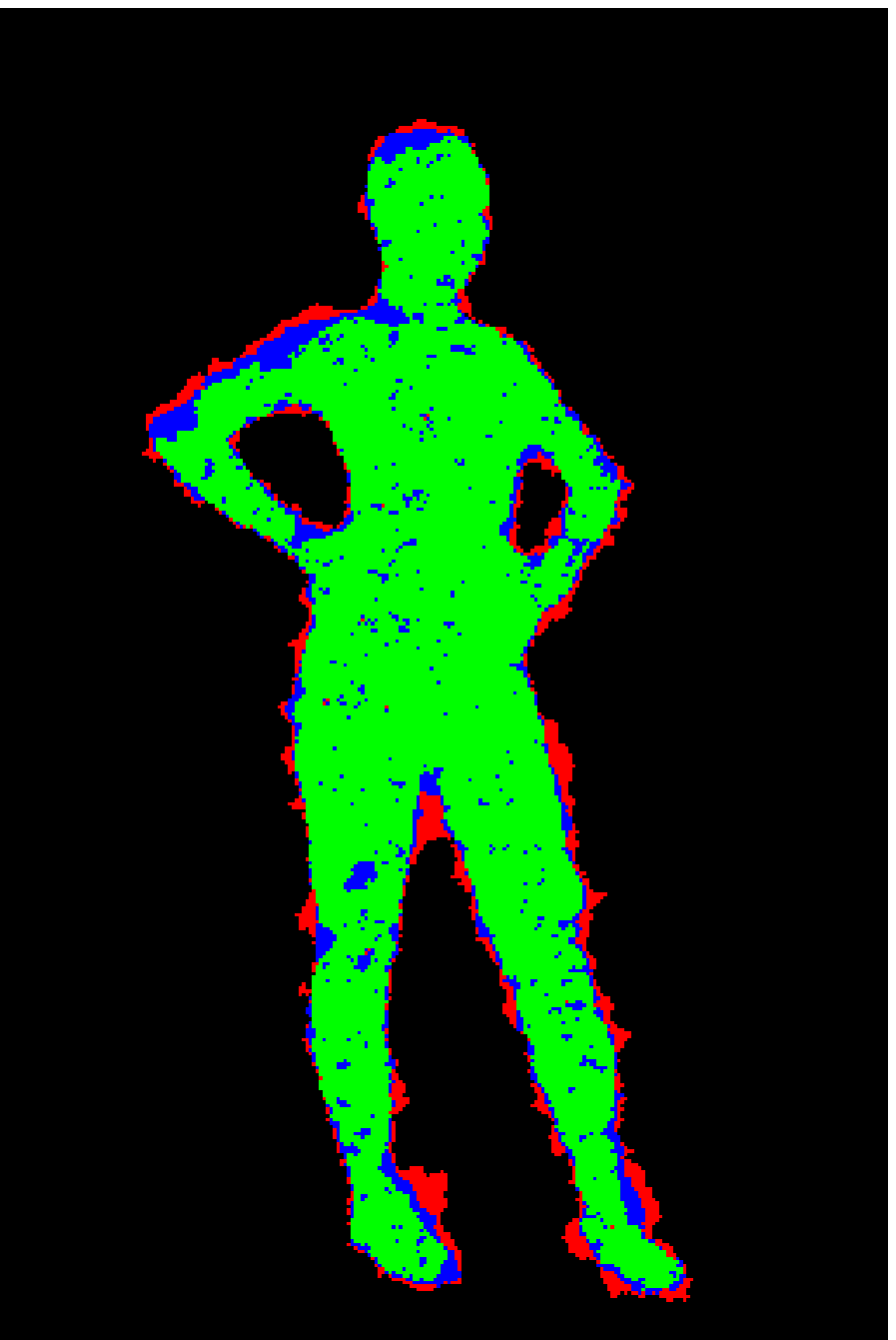}\\%
		\includegraphics[width=\linewidth]{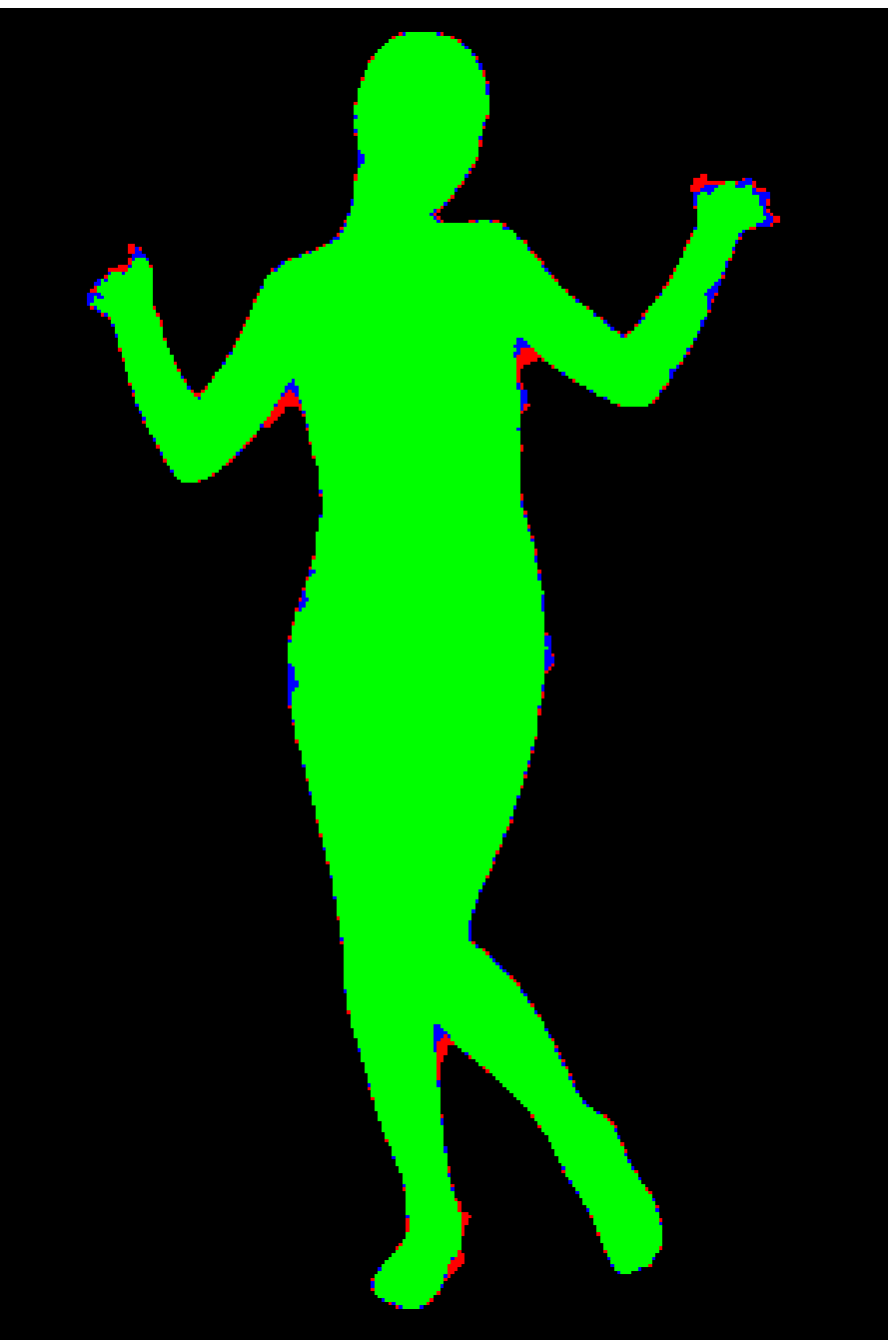}\\
		\footnotesize CCP
	\end{minipage}
	\hfill
	\vspace{\vspacecapt}
	\caption{Example images illustrating the utility of topology priors.}
	\label{fig:person}
\end{figure}

The CCP constraint is particularly useful when an unknown number of
handles exist (e.g. deformable objects). Objects with a known number
of handles in 3D projected onto a 2D plane can have any number of
handles. We show two example images of a human and the resulting
thresholded histogram image in Figure \ref{fig:person}. In the first image,
the handles formed by the arms are not captured well with TP and
GP. In the second image, the vignetting causes the
unconstrained and the GP to group some background with 
foreground.

%% file: conclusion.tex
We have presented a new MCMC sampler for implicit shapes. We have shown how to design a proposal such that every proposed sample is accepted. Unlike previous methods, GIMH-SS efficiently samples PDE-based and graph-based energy functionals and does not require the evaluation of the gradient of the energy functional (which may be hard to compute). Additionally, GIMH-SS was extended to include hard topological constraints by only proposing samples that are topologically consistent with some prior.